\documentclass[a4paper]{article}
\usepackage[utf8]{inputenc}
\usepackage[english]{babel}
\usepackage{amsmath} 
\usepackage{graphicx}
\usepackage{fancyhdr}
\usepackage{setspace}
\usepackage{lineno}	 
\usepackage{lastpage}
\usepackage[margin=1in]{geometry}
\usepackage{caption}  
\usepackage{amsmath,amssymb}
\usepackage{multirow}
{
	\fancyhead[L]{\myshorttitle}
	\fancyhead[R]{\small{\textit{\the\day.\the\month.\the\year}}}
	\fancyfoot[L]{\copyright \small{\textit{\myauthorshort}}}
	\fancyfoot[R]{Page \thepage\ of \pageref*{LastPage}}
	\cfoot{}
}

\fancypagestyle{firststyle}
{
	\fancyhead[L]{\copyright \small{\textit{\myauthorshort}}}
	\fancyhead[C]{}
	\fancyhead[R]{Page \thepage\ of \pageref*{LastPage}}
	\fancyfoot[L]{\footnotesize \rule{0.25\textwidth}{0.4pt} \newline Corresponding author: $^*$A. M. Schweidtmann, E-mail: a.schweidtmann@tudelft.nl}
	\fancyfoot[C]{}
	\fancyfoot[R]{}
}

\renewenvironment{abstract}{\noindent\textbf{Abstract:}}{}

\usepackage{authblk}
\usepackage{polski}
\usepackage{url}
\usepackage{algorithm}
\usepackage{algpseudocode}
\usepackage[version=4]{mhchem}
\usepackage{siunitx}
\usepackage{subcaption}
\usepackage{booktabs}
\usepackage{tabulary}
\usepackage{supertabular}
\usepackage{eurosym}
\usepackage{csquotes}
\usepackage[usenames,dvipsnames]{xcolor}
\usepackage[normalem]{ulem}
\algdef{SE}{Begin}{End}{\textbf{begin}}{\textbf{end}}
\DeclareSIUnit\year{a}
\newcolumntype{P}[1]{>{\centering\arraybackslash}p{#1}}
\newcolumntype{M}[1]{>{\centering\arraybackslash}m{#1}}
\newcolumntype{N}[1]{>{\raggedright\arraybackslash}m{#1}}

\newcommand{\mytitle}{Deep reinforcement learning for process design: Review and perspective}
\newcommand{\myshorttitle}{Reinforcement learning in process design}
\newcommand{\myauthor}{Qinghe Gao$^{1}$, Artur M. Schweidtmann$^{1,*}$} 
\newcommand{\myauthorshort}{Q. Gao and A. M. Schweidtmann}
\author{\myauthor}

\usepackage[colorlinks,linkcolor=blue,citecolor=blue,urlcolor=blue,pdftitle={\mytitle},pdfauthor={Artur M. Schweidtmann}]{hyperref} 
\usepackage{cleveref}

\begin{document}
\thispagestyle{firststyle}
	\begin{flushleft}\begin{large}\textbf{\mytitle}\end{large} \end{flushleft}
	\myauthor 
	
	\begin{flushleft}\begin{small}
			$^1$ Delft University of Technology, 
			Department of Chemical Engineering, 
			Van der Maasweg 9, 
			Delft 2629 HZ, 
			The Netherlands\\[0.15cm]
		\end{small}
	\end{flushleft}
\begin{abstract}
\noindent
The transformation towards renewable energy and feedstock supply in the chemical industry requires new conceptual process design approaches.
Recently, deep reinforcement learning, a subclass of machine learning, has shown the potential to solve complex decision-making problems and aid sustainable process design. However, its suitability in static process design still needs to be examined. 
We discuss the advantages and disadvantages of reinforcement learning for process design. 
Then, we survey state-of-the-art research through three major elements: (i) information representation, (ii) agent architecture, and (iii) environment and reward. 
Moreover, we discuss perspectives on underlying challenges and promising future works to unfold the full potential of reinforcement learning for process design in chemical engineering.
\end{abstract}


\section{Introduction}
\label{2023RLreview:introduction}

The chemical industry is facing a rapid paradigm shift towards a circular economy based on renewable energy and feedstock supply~\cite{MeramoHurtado2021_Processsynthesisanalysis, MartinezHernandez2017_Trendssustainableprocess}. 
This poses several challenges for conceptual process design due to the increasing complexity of the design task, the lack of experienced engineers, and the pressure on improving sustainability and profitability while shortening development times.  
Thus, there is a need for new methodologies and tools that support engineers to design sustainable processes in a more efficient way. 

\begin{figure}
    \centering
    \includegraphics[width=\textwidth]{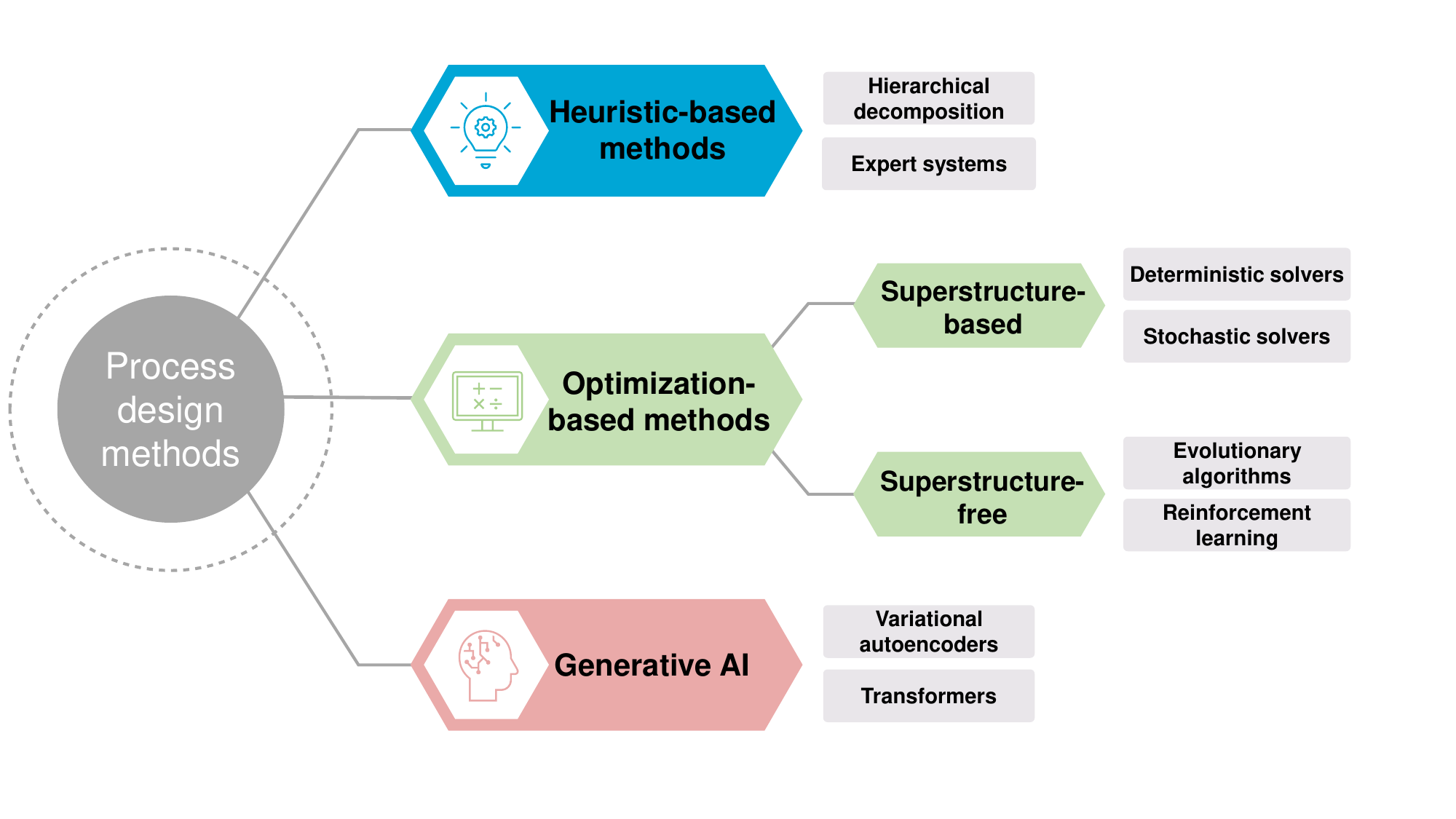}
    \caption{Overview of computer-aided process design methods.}
    \label{fig:PD_methods}
\end{figure}

Computer-aid process design (CAPD) is widely used in process systems engineering (PSE) for conceptual process design~\cite{Umeda1983_Computeraidedprocess, Dimian2008_Chemicalprocessdesign}, which can be classified into three methodologies as illustrated in Figure~\ref{fig:PD_methods}: (i) heuristic-based methods; (ii) optimization-based methods; and (iii) the emerging field of generative artificial intelligence (AI).
Heuristic-based methods rely on a set of rules derived from experience, insights, and engineering knowledge, making them the most commonly used approaches due to their ease of application.
Optimization-based approaches are commonly used to identify optimal design and operating variables for given process structures. 
Particularly, derivative-based optimization algorithms are already available in commercial process simulation software to determine those.
To determine optimal process structures, superstructure-based methods are the de facto state of the art, where possible process alternatives are modeled and subsequently solved by mixed-integer nonlinear optimization (MINLP) methods~\cite{Yee1990_Simultaneousoptimizationmodels, Mencarelli2020_reviewsuperstructureoptimization}. 
Within this paradigm, two solver types are typically utilized:  deterministic (e.g., branch and bound algorithms like BARON or MAiNGO), and stochastic solvers (e.g., genetic algorithms).  
While superstructure methods have been very successful in PSE, they also have many shortcomings that limit industrial applications~\cite{Mitsos2018_Challengesprocessoptimization}, including (manual) setup of all process alternatives, the need to implement process models in an optimization environment, and the difficulties of solving resulting MINLPs. 
Superstructure-free methods dispense the need for predefined superstructures. 
For example, evolutionary algorithms, typically adopt a two-level decomposition approach~\cite{Mencarelli2020_reviewsuperstructureoptimization}. 
First, they generate alternative flowsheets which are then evaluated through an optimization algorithm.
Finally, a few very recent works proposed the use of generative AI methods for the generation of process structures~\cite{Hirtreiter2023_automaticgenerationcontrol, nabil2022generative}. 
However, those works require large training data and are not the focus of this review.

Recently, deep reinforcement learning (RL) has shown its potential to solve complex sequential dynamic decision-making problems at human-like or even superhuman level~\cite{Mnih2013_PlayingAtariDeep, Kempka2016_ViZDoomDoombased, Silver2018_generalreinforcementlearning}.
RL is a computational approach for goal-directed learning and decision-making through the direct interaction of an agent with its environment~\cite{Sutton2018_Reinforcementlearningintroduction}.
RL is primarily developed to solve discrete-time dynamic optimization problems formulated as Markov Decision Processes. 
Consequently, RL is based on the Bellman optimality equation, which is similar to the Hamilton-Jacobi-Bellman (HJB) equation and Pontryagin's Maximum Principle (PMP) for continuous state and action spaces in the control theory.
Also, RL has seen its first applications in chemical engineering for process control~\cite{Hoskins1992_Processcontrolvia, Spielberg2017_Deepreinforcementlearning} and scheduling~\cite{Hubbs2020_deepreinforcementlearning, Lee2022_Deepreinforcementlearning}. 
Notably, Yokogawa is even using RL to operate an industrial chemical process since 2022.

This review and perspective paper aims to provide a critical examination of the application of RL for process design.
In the context of process design, RL can be considered a superstructure-free and model-free method, which iteratively places unit operations with corresponding design and operating variables.
It evaluates the resulting flowsheets at every iteration and aims to maximize the objective functions.
In the recent literature, there have been a few first steps towards applying RL as a static optimization algorithm for stationary process design including absorption–stripping process~\cite{Chen2014_CostreductionCO2}, energy systems design~\cite{Perera2020_Introducingreinforcementlearning, Caputo2023_Designplanningflexible}, unit operation design~\cite{Sachio2022_Integratingprocessdesign}, separation process~\cite{Midgley2020_DeepReinforcementLearning, Kalmthout2022_Synthesisseparationprocesses, Goettl2021_Automatedsynthesissteady, Goettl2021_AutomatedFlowsheetSynthesis, Goettl2022_UsingReinforcementLearning, Khan2020_Searchingoptimalprocess, Khan2022_Designingprocessdesigner, Seidenberg2023_Boostingautonomousprocess}, solvent extraction process design~\cite{Plathottam2021_Solventextractionprocess}, single mixed-refrigerant process design~\cite{Kim2023_Processdesignoptimization}, and synthesis reaction process design~\cite{Wang2022_Reinforcementlearningautomated, Stops2022_Flowsheetgenerationhierarchical, Gao2023_Transferlearningprocess}.

The use of RL for stationary process design is controversial in scientific discussions.
Most notably, RL is better suited for dynamic, sequential-decision problems rather than static ones. 
Also, clear comparisons and benchmarks between RL and other design methods are lacking in previous literature. 
Therefore, at the moment, the suitability of RL for process design is questionable and remains an open question. 
Here, we present the main differences between RL and existing superstructure-free methods in the context of process design: 
\begin{itemize}
    \item \textbf{Computational efficiency (static vs. dynamic optimization)} - RL is better suited for dynamic, sequential-decision problems rather than static ones. However, prior studies have applied RL to stationary process design, constituting static optimization problems.
    This may drastically increase complexity and decrease computational efficiency (called "sample efficiency" in RL). As a side note, the capability of RL in dynamic optimization paves the way for solving integrated design and operation problems (see Section~\ref{2023RLreview:pers_control_design}).
    \item \textbf{Iterative build-up vs. full flowsheet generation} - RL sequentially generates flowsheets (unit by unit), differing from evolutionary superstructure-free methods that construct flowsheets as a whole.
    While this sequential build-up increases computational complexity, there are also potential advantages.
    Generating feasible flowsheets and simulating them is challenging, often causing convergence problems in evolutionary superstructure-free methods. 
    In contrast, the iterative strategy of RL promotes convergence and intermediate simulations of incomplete flowsheets may provide valuable information for learning. 
    However, this additional information comes at the cost of additional simulation time. 
    \item \textbf{Inference (online) vs. optimization (offline)} - A significant distinction between RL and other optimization methods lies in the solution time for similar, recurring problems. 
    RL involves training a policy once, which is then utilized during inference to rapidly predict near-optimal solutions, offering a significant advantage in time-sensitive control applications.
    In contrast, classical optimization algorithms solve problems individually, typically requiring long runtimes for each problem instance. 
    In process design, long optimization times are usually not an issue (unless it becomes intractable in the case of large non-convex MINLPs). 
    Thus, classical optimization methods are usually well-suited. 
    However, the rapid inference capability of RL may also provide new opportunities for process design. For example, RL agents might be integrated into flowsheet simulation software to automatically and immediately suggest near-optimal design options to users. Also, fast solutions of design problems can be advantageous when a large numbers of design problems need to be solved (e.g., as subproblems in larger optimization studies). 
    \item \textbf{Learning capacity} - RL possesses a substantially greater learning capacity compared to standard evolutionary methods (e.g., more trainable parameters). 
    For instance, state-of-the-art deep RL algorithms can incorporate extensive networks of learnable parameters, potentially exceeding billions of parameters. 
    This enhanced learning potential facilitates inference and allows for retaining information, unlike genetic algorithms which typically lose details about previous populations.
    This substantial learning capacity of RL holds the potential for learning more complex dependencies between design actions and results. 
    However, the high learning capacity also presents severe drawbacks such as a vast amount of training data and extensive training duration (measured in epochs within RL). 
\end{itemize}

When comparing RL with standard optimization methods, its most notable advantages include larger learning capacity and inference ability. However, at the same time, RL typically demands significantly more training simulations than static optimization solvers, which leaves doubt on whether its potential advantages outweigh the disadvantages in the context of process design.
Current literature applying RL to process design neglects its inference capabilities, learning capacity, and dynamic optimization capabilities, predominantly utilizing the training phase of RL as an evolutionary optimization strategy for static problems. 
Thereby, they essentially merge the drawbacks of both worlds. 
Additionally, there is a lack of computational comparison between RL and traditional process design methods.
In the following, we critically examine the existing literature on RL in process design (Section~\ref{2023RLreview:state_of_the_art}) and highlight future perspectives (Section~\ref{2023RLreview:Perspectives}).

\section{State of the art}
\label{2023RLreview:state_of_the_art}
\begin{figure}
    \centering
    \includegraphics[width=\textwidth]{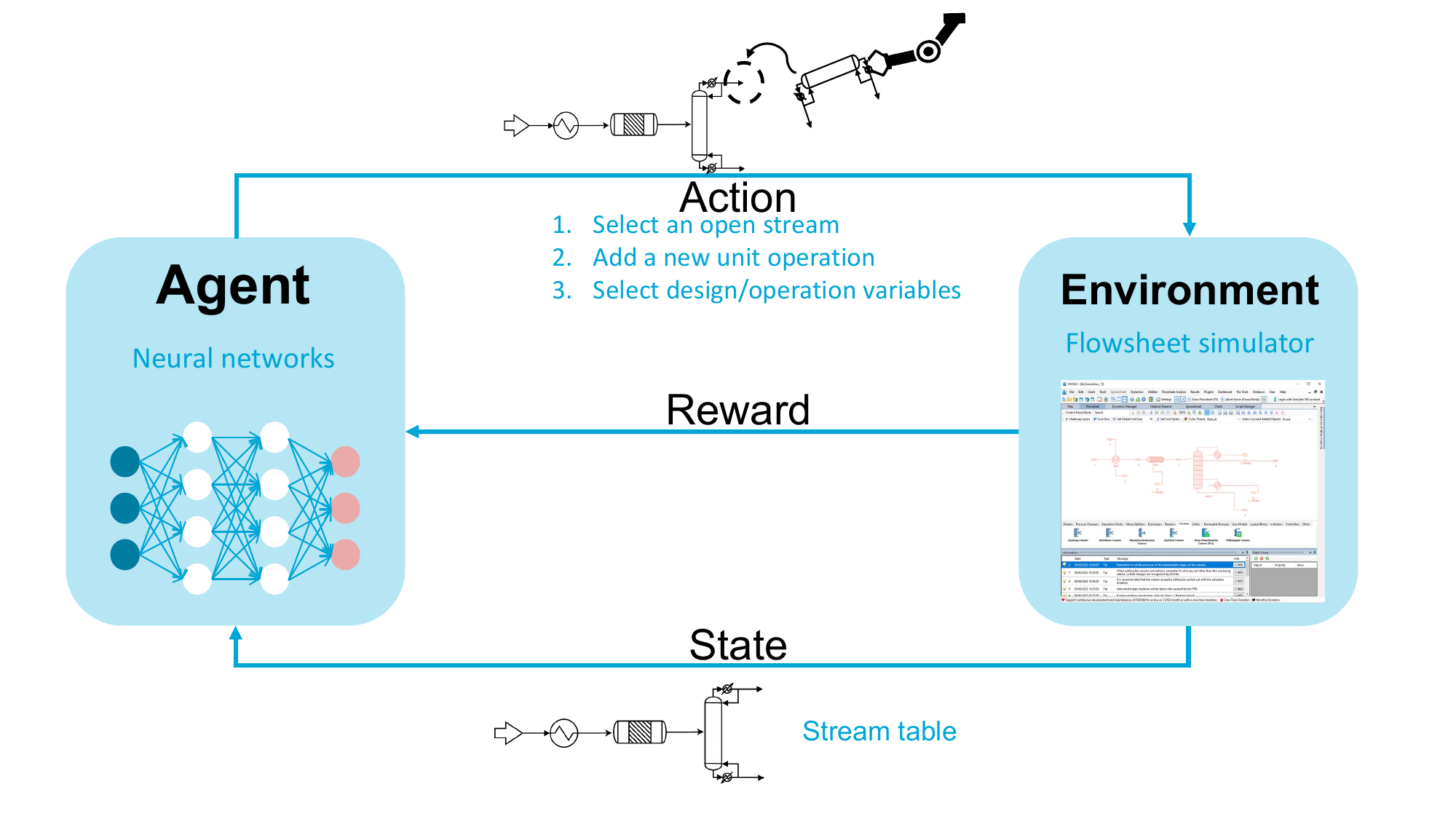}
    \caption{ General framework of reinforcement learning for process design.}
    \label{fig:RL_framework}
\end{figure}

The general framework of RL in process design is shown in Figure \ref{fig:RL_framework}. 
The agent learns to design processes by iteratively placing unit operations with design and operating variables, and simulating the resulting processes in the environment, ultimately obtaining the optimal policy $\pi^*$ which designs optimal processes.
Mathematically, this problem can be formulated as Markov decision processes (MDP): $M = \{S, A, T, R\}$ with states $\mathbf{s} \in S $, actions $\mathbf{a} \in A $, the transition function $T: S  \times A \times S  \to \left[0,1\right]$, and the reward function $R: S \times A\times S  \to \mathbb{R}$. 
In the context of process design, the states $\mathbf{s}$ represent the flowsheet topology as well as all relevant design specifications, operating variables, thermodynamic stream data, flowrates, and compositions.
The agent takes the current states $\mathbf{s}$ as input to take actions $\mathbf{a}$. These actions can include design and operating variables. In chemical processes, design variables are usually determined during the initial design phase and typically remain fixed throughout the operation, such as equipment size. Operating variables can be adjusted during operation (e.g., flow rates, and pressures). Furthermore, the actions contain discrete choices (e.g., the selection of open streams, unit operation types, or number of stages) as well as continuous choices (e.g., the length of a reactor or operating flowrates), namely hybrid action space.
Usually, decisions in process design are also hierarchical.
For example, the agent first determines an open stream to add a unit operation, then the type of unit operation, then design variables, and finally operating variables. 
After a new unit operation is added, the new flowsheet is simulated in an environment (e.g., a process simulation software). 
After finishing a flowsheet, a numerical reward $\mathbb{R}$ is returned to the agent. 
This corresponds to the objective for optimization. 
By repeating the design of multiple flowsheets and receiving corresponding rewards, the agent learns to design processes that maximize the reward.

In this section, we survey the RL state-of-the-art literature (summarized in Table~\ref{tab:literature_table}) based on (i) information representation, (ii) agent architecture, and (iii) environment and reward.

\begin{table}[htbp]
    \centering
     \caption{An overview of the reviewed literature and different choices of elements in RL for process design. We use Repr. to indicate information representation. Furthermore, we utilize Dis., Cont., and Hier. to denote discrete, continuous, and hierarchical action space of RL. Within the decisions of RL, we use Topo., Des. and Oper. to represent actions involved in changing flowsheet topologies, selecting design variables and operating variables, respectively.}
    \resizebox{\linewidth}{!}
    {%
    \begin{tabular}{lllllllllll}
        \toprule
        \multicolumn{1}{c}{\multirow{2}[3]{*}{\shortstack[c]{Ref.}}} & \multicolumn{1}{c}{\multirow{2}[3]{*}{\shortstack[c]{Repr.}}} & \multicolumn{1}{c}{\multirow{2}[3]{*}{\shortstack[c]{Agent architecture}}}  & \multicolumn{1}{c}{\multirow{2}[3]{*}{\shortstack[c]{Environment}}}  &\multicolumn{3}{c}{Action space}  & \multicolumn{3}{c}{Decisions}   \\
         \cmidrule(lr){5-7}\cmidrule(lr){8-10}
        \multicolumn{1}{c}{}   & \multicolumn{1}{c}{} & \multicolumn{1}{c}{} & \multicolumn{1}{c}{}& \multicolumn{1}{c}{Dis.}& \multicolumn{1}{c}{Cont.}& \multicolumn{1}{c}{Hier.}  & \multicolumn{1}{c}{Topo.}& \multicolumn{1}{c}{Des.} & \multicolumn{1}{c}{Oper.} \\
        \midrule
        \cite{Midgley2020_DeepReinforcementLearning}  & Matrix & Actor-Critic (SAC) & COCO & \checkmark &\checkmark  &&\checkmark &\checkmark &\checkmark  \\
        \cite{Kalmthout2022_Synthesisseparationprocesses}& Matrix & Actor-Critic (SAC) & Aspen Plus & \checkmark &\checkmark & &&\checkmark &\checkmark \\
        \cite{Chen2014_CostreductionCO2}&Matrix & Actor-Critic ( - )  & Aspen Plus & &\checkmark&& & &\checkmark\\
        \cite{Plathottam2021_Solventextractionprocess}& Matrix & Actor-Critic (PPO) & Short-cut & &\checkmark&& & \checkmark & \\
        \cite{Goettl2021_Automatedsynthesissteady} \cite{Goettl2021_AutomatedFlowsheetSynthesis} \cite{Goettl2022_UsingReinforcementLearning} & Matrix & Actor-Critic (Two players) & Short-cut &\checkmark&& \checkmark&\checkmark &&  \\      
        \cite{Khan2022_Designingprocessdesigner}\cite{Seidenberg2023_Boostingautonomousprocess}& Matrix & Actor-Critic (PPO) & Short-cut &  \checkmark &\checkmark  &\checkmark  &\checkmark &\checkmark &\checkmark  \\  
        \cite{Caputo2023_Designplanningflexible} & Matrix & Actor-Critic (ACER) &Short-cut & \checkmark &&&&\checkmark& \\
        \cite{Perera2020_Introducingreinforcementlearning} & Matrix & Policy-based (Policy search) & Short-cut & &\checkmark& &&&\checkmark\\
        \cite{Sachio2022_Integratingprocessdesign}& Matrix & Policy-based (PG) & Short-cut & &\checkmark& &&\checkmark&\checkmark \\
        \cite{Wang2022_Reinforcementlearningautomated}  & Matrix & Value-based (DQN) & IDAES &\checkmark &&&\checkmark&& \\
        \cite{Kim2023_Processdesignoptimization}&Matrix & Value-based (DQN) & UniSim & &\checkmark&&&\checkmark&\checkmark\\
        \cite{Khan2020_Searchingoptimalprocess}  & Matrix & Value-based (Q-learning) &Short-cut &  \checkmark &\checkmark & &\checkmark&\checkmark&\\
        \cite{Stops2022_Flowsheetgenerationhierarchical}& Graph & Actor-Critic (PPO)  &Short-cut &  \checkmark &\checkmark  &\checkmark&\checkmark &\checkmark &\checkmark \\
        \cite{Gao2023_Transferlearningprocess}& Graph & Actor-Critic (PPO)&DWSIM &  \checkmark &\checkmark &\checkmark  &\checkmark &\checkmark &\checkmark\\
        \bottomrule
    \end{tabular}
    }
    \label{tab:literature_table}
\end{table}

\subsection{Information representation}
\label{2023RLreview:sub-data representation}
Chemical processes comprise various information, such as process topology, thermodynamic states, flowrates, concentrations, design variables, operating variables, components, and underlying mechanistic knowledge. 
The meaningful representation of the chemical process information is critical for the learning and generalization of RL agents.  
For RL in process design, there are currently two methods for information representations: Matrix~\cite{Midgley2020_DeepReinforcementLearning, Perera2020_Introducingreinforcementlearning, Kalmthout2022_Synthesisseparationprocesses, Plathottam2021_Solventextractionprocess,   Goettl2021_Automatedsynthesissteady, Goettl2021_AutomatedFlowsheetSynthesis,Goettl2022_UsingReinforcementLearning,Caputo2023_Designplanningflexible,Seidenberg2023_Boostingautonomousprocess,Sachio2022_Integratingprocessdesign,Kim2023_Processdesignoptimization, Chen2014_CostreductionCO2, Wang2022_Reinforcementlearningautomated, Khan2020_Searchingoptimalprocess, Khan2022_Designingprocessdesigner} and graph~\cite{Stops2022_Flowsheetgenerationhierarchical, Gao2023_Transferlearningprocess}.

In matrix-based representation, flowsheets are represented by fixed-size matrices. 
Within the flowsheet matrix, the connectivity, stream compositions, thermodynamic stream data, and design variables are usually concatenated. 
For example, G\"ottl et al.~\cite{Goettl2021_Automatedsynthesissteady, Goettl2021_AutomatedFlowsheetSynthesis,Goettl2022_UsingReinforcementLearning} represented flowsheets as $16\times28$ matrices, where each row represents a stream and encompasses four parts: \{$\mathbf{v}, \mathbf{u},\mathbf{d},\mathbf{t}$\}. 
$\mathbf{v}_i$ has five entries, which describe the molar fractions and total molar flowrate of stream $i$. 
$\mathbf{u}_i$ stores the type of the unit operation that is downstream of stream $i$ as one-hot encoding. 
Furthermore, $\mathbf{d}_i$ stores the connectivity of unit operations and has sixteen entries (i.e., this corresponds to the adjacency matrix). 
Finally, $\mathbf{t}_i$ has two entries: the first entry indicates whether the task is terminated (0 if not terminated), and the second entry indicates whether stream $i$ is still unused (0 if unused).
Most of the previous publications use matrix representations of flowsheet states~(c.f. Table~\ref{tab:literature_table}). 

In graph-based representation, flowsheets are represented by directed heterogeneous graphs. 
Flowsheet graphs consist of nodes and edges. Unit operations are represented by nodes, also referred to as vertices $v \in V $, and streams are represented by edges $e_{vw} \in E$ connecting two nodes $v$ and $w$.
Importantly, node feature vector $\mathbf{f}^{v} \in F^{V} $, and edge feature vector $\mathbf{f}^{e_{vw}} \in F^{E} $ are associated with each node and edge, respectively. 
Within the node feature vectors, types of unit operation, design specifications, and operating points are encoded. 
The edge feature vectors contain thermodynamic states, concentrations, and flowrates. 
In the past, only our previous works used graph representations of flowsheets for RL~\cite{Stops2022_Flowsheetgenerationhierarchical, Gao2023_Transferlearningprocess}.

The comparison between flowsheet matrices and flowsheet graphs is still an open research question in the context of RL in process design. 
Flowsheet matrices are easier to implement than flowsheet graphs and are used by the majority of the literature as shown in Table~\ref{tab:literature_table}. 
Flowsheet matrices are processed by RL agents using multilayer perceptrons~(MLPs) or convolutional neural networks~(CNNs). 
However, every flowsheet graph has $N!$ different adjacency matrices. 
CNNs and MLPs are not permutation equivariance
\begin{equation}
    f(\mathbf{P}^T\mathbf{xP}) \neq \mathbf{P}^Tf(\mathbf{x})\mathbf{P} \notag
\end{equation}
where $\mathbf{P}$ is a permutation matrix, $\mathbf{x}$ is the input matrix and $f$ is a MLP or CNN~\cite{Hamilton2020_GraphRepresentationLearning}. 
This means that such models depend on the arbitrary order of rows/columns in the flowsheet matrix and thus, cannot generalize over flowsheet topologies. 
Also, the neighborhood of an entry in the matrix does not correspond to physical connectivity which makes learning using MLPs/CNNs more difficult as it requires learning long-range interactions. 
In contrast, (message-passing) graph convolutional networks (GCNs) are permutation equivariance and they learn from the actual connectivity of flowsheet graphs. 
Furthermore, MLPs and CNNs require fixed-size inputs, e.g., a pre-defined maximum number of unit operations and streams, while GCNs are size-independent~\cite{Zhou2020_Graphneuralnetworks}. 

\subsection{Agent architecture}
\label{2023RLreview:sub-agent architecture}
RL agents consist of two components: A policy and a learning algorithm. 
The policy describes the behavior of the agent, mapping the current state $\mathbf{s}$ into an action $\mathbf{a}$: $\pi(\mathbf{s}) = \mathbf{a}$.
It is parameterized by function approximators such as MLPs.
The learning algorithm is used to continuously update the policy based on the actions, states, and rewards. 
Depending on the learning algorithms, the agent can be characterized into three types: Value-based, policy-based,  and actor-critic-based (AC).

Value-based agents learn a functional approximator of the value function ($V_{\pi}(\mathbf{s})$) to take actions.
The value function outputs the expected returns after the current process step $t$ given a state $\mathbf{s}$ and  a policy $\pi$: $V_{\pi}(\mathbf{s}) = \mathbb{E}_{\pi} \left [ G_{t} |\mathbf{s}  \right ]$, where returns $G_{t}$:
\begin{equation}
    G_{t} = R_{t+1}+ \gamma R_{t+2} +\dots+ \gamma^k R_{t+k+1} = \sum_{k=0}^{\infty} \gamma^k R_{t+k+1} 
    \label{eq: expected_return}
\end{equation}
where $\gamma^k$ are the discount rates to determine the present value of future rewards, and $k$ is the process step from $t$ to the end of the episode.
Similarly, we can also derive the state-action value function, namely the quality function $Q_{\pi}$, which calculates the expected returns given a state $\mathbf{s}$ and action $\mathbf{a}$, following policy $\pi$: $Q_{\pi}(\mathbf{s},\mathbf{a}) = \mathbb{E}_{\pi} \left [  G_{t} |\mathbf{s},\mathbf{a}  \right ]$. Depending on the calculated V-/Q-value, different search algorithms such as best-first search or nearest neighbors are used to choose the final action. 
In the context of RL in process design, three works~\cite{Khan2020_Searchingoptimalprocess,Wang2022_Reinforcementlearningautomated,Kim2023_Processdesignoptimization} deployed Q-learning based agent to perform process synthesis tasks.
However, traditional value-based agents can only take discrete actions, which hinders further development because continuous decision-making of design or operating variables is vital in process design tasks.

Policy-based agents directly learn a functional approximator of the policy function. 
Specifically, the policy approximator $\pi_{\theta}$ maps the current states $\mathbf{s}$ to the actions $\mathbf{a}$: $\pi_{\theta}(\mathbf{s}) = \mathbf{a}$. 
And the optimal policy $\pi^{*}$ is obtained by maximizing the expected return $\mathbb{E}_{\theta} \left [ G_{t} \right ]$ through policy gradient or policy-search methods.  
In the context of RL in process design, Sachio et al.~\cite{Sachio2022_Integratingprocessdesign} and Perera et al.~\cite{Perera2020_Introducingreinforcementlearning} utilized policy gradient methods and policy-search methods to perform process design tasks, respectively. 
Compared to the value-based approach, the policy-based agent can handle both discrete and continuous actions.
However, policy-based methods are known for high variance and sub-optimal local solutions~\cite{Nachum2017_BridgingGapValue}.

AC agents combine the advantages of value-based and policy-based methods.
AC consists of an actor, working as a functional approximator of the policy function, and a critic, serving as a functional approximator of the value function. 
Therefore, AC agents explicitly optimize both value and policy functions and are able to process both discrete and continuous action spaces. 
In the context of RL in process design,  different types of AC agents have been used such as Proximal Policy Optimization (PPO)~\cite{ Seidenberg2023_Boostingautonomousprocess}~\cite{Plathottam2021_Solventextractionprocess}~\cite{Stops2022_Flowsheetgenerationhierarchical}~\cite{Gao2023_Transferlearningprocess}, Soft Actor-Critic (SAC)~\cite{Midgley2020_DeepReinforcementLearning,Kalmthout2022_Synthesisseparationprocesses}, Two-player game~\cite{Goettl2021_Automatedsynthesissteady, Goettl2021_AutomatedFlowsheetSynthesis, Goettl2022_UsingReinforcementLearning}, and Sample Efficient Actor Critic with Prioritized Experience Replay (ACER)~\cite{Caputo2023_Designplanningflexible}.

The choice of agent architecture for RL in process design is an open question.
AC RL is deemed to be a viable option because it combines the advantages of value-based and policy-based and can handle both discrete and continuous decisions.
Specifically,  PPO is the most popular algorithm in process design tasks with the advantage of less complicated implementation and a stable learning process.
However, PPO is an on-policy algorithm which means the optimized policy is the same as the policy for action selection.
Therefore, PPO is less data-efficient than off-policy algorithms, such as SAC and ACER, which may take less time and fewer training episodes.
Moreover, AC RL comes with challenges, including complex implementation, computational demands when optimizing both actor and critic networks concurrently, and potential convergence issues~\cite{Nachum2017_BridgingGapValue, Sutton2018_Reinforcementlearningintroduction}.

\subsection{Environment and reward}
\label{2023RLreview:sub-environment_and_reward}
The environment simulates the processes and computes a reward as feedback to the agent.
Selecting an appropriate accuracy level for the environment is a vital task for RL in process design and depends on the task-specific requirements and modeling intent.
There are two main levels of accuracy: Shortcut and rigorous simulators.
Shortcut simulators utilize approximated process models to ensure tractability but can be inaccurate.
Rigorous simulators involve more accurate process models that require longer computation times, as previous studies indicated~\cite{Gao2023_Transferlearningprocess,Kalmthout2022_Synthesisseparationprocesses}.
In the past, RL for process design has used multiple process simulation software including open-source (DWSIM, IDEAS), non-commercial (COCO), and commercial (UniSim, Aspen Plus) alternatives. 
Additionally, Seidenberg et al.~\cite{Seidenberg2023_Boostingautonomousprocess} leveraged knowledge graphs to retrieve information about the design task, process knowledge, and the current state of the process. 
Notably, this knowledge graph was part of a manually-implemented environment and not directly accessible to the RL agent. 
Thus, the agent also relied on a flowsheet matrix representation as states. 

Previous research optimized towards a single economic objective~\cite{Midgley2020_DeepReinforcementLearning, Kalmthout2022_Synthesisseparationprocesses, Seidenberg2023_Boostingautonomousprocess, Goettl2021_Automatedsynthesissteady,Goettl2021_AutomatedFlowsheetSynthesis,Goettl2022_UsingReinforcementLearning, Caputo2023_Designplanningflexible, Stops2022_Flowsheetgenerationhierarchical, Gao2023_Transferlearningprocess, Khan2020_Searchingoptimalprocess,Sachio2022_Integratingprocessdesign, Chen2014_CostreductionCO2}.
Also, some works integrate purity, recovery, power consumption, and product flow rate into scalar reward functions~\cite{Plathottam2021_Solventextractionprocess, Wang2022_Reinforcementlearningautomated, Kim2023_Processdesignoptimization}.

\section{Perspectives}
\label{2023RLreview:Perspectives}

Despite the first demonstrations of RL for process design, it is still unclear if RL outperforms existing design methods. 
In our view, the big research challenge is the generalization of RL models and the use of its inference capabilities. 
The training phase of current RL frameworks is essentially used like a derivative-free optimization approach (e.g., a genetic algorithm) to optimize the process topology for one particular case study. 
Thus, a re-training is needed for a new case study and the agent fails to transfer its learning to new situations. 
In general, deep RL has an inference and a high learning capacity. 
The derivative-free optimization approach with RL does not use the full potential of RL. 
However, useful application of inference requires generalization across multiple case studies. 
This generalization requires an extension of the information representation and agent architecture to account for process-relevant knowledge. 
This includes domain expertise, prior process data, and physical constraints which are typically employed by engineers when designing chemical processes. 
Integrating this information would allow the RL agent to see what "drives the process" and ultimately unlock the full potential of RL by learning from multiple processes. 
We envision that RL will generalize (to some extent) and ultimately design processes at inference time. 
In this section, we provide our perspectives on underlying challenges and a number of other promising future works.

\subsection{Information representation}
\label{2023RLreview:pers_information_representations}

Information representation is critical for RL since it encapsulates the current state of the environment, which directly affects decision-making for agents. 
However, current information representations still lack mechanistic knowledge and relevant process information.  
Furthermore, it neglects the appropriate representation of molecules.
Integrating the above information will significantly benefit the RL agent in generalization.
Numerous representation methods could be potentially incorporated into RL in process design for process and molecular information representation.
For example, Simplified Flowsheet Input-Line Entry-System (SFILES)~\cite{dAnterroches2005_ProcessFlowsheetGeneration}, SFILES 2.0~\cite{Vogel2023_SFILES2.0extended}, eSFILES~\cite{Mann2023_IntelligentProcessFlowsheet}, and knowledge graphs~\cite{Seidenberg2023_Boostingautonomousprocess} could be used to enhance process representations.
Similarly, molecular descriptors~\cite{Todeschini2010_Moleculardescriptors}, molecular graphs~\cite{Schweidtmann2020_GraphNeuralNetworks}, SMILES~\cite{Weininger1988_SMILESchemicallanguage}, knowledge graphs~\cite{Fang2022_MolecularContrastiveLearning}, and hypergraphs~\cite{Kajino2018_MolecularHypergraphGrammar}
could be used to encode molecular information.

\subsection{Agent architecture}
In this section, we identify the limitations and potential improvements of the current agent architecture.
\subsubsection{Integration of mechanistic knowledge}
Current RL algorithms are not sufficient to transfer knowledge into the new processes because RL agents have a limited understanding of mechanistic knowledge and physical properties.
Future work could consider implementing a physics-informed RL agent by encoding information-rich representations such as knowledge graphs or hypergraphs to inform the agent.
Furthermore, fundamental concepts, such as thermodynamic driving forces (Gibb's free energy), could be included in the RL agents. 
This allows the agent to learn general concepts that can be translated into other problems because they are based on physics.

\subsubsection{Integration of prior data}
\label{2023RLreview:pers_prior_data}
RL for process design is currently initialized randomly, which can lead to suboptimal solutions, excessive training times, and frequent convergence issues. 
Meanwhile, there is a large number of existing digitized chemical process data from simulation files and images~\cite{Balhorn2022_FlowsheetRecognitionusing}, which can potentially accelerate the learning process of the agent. 
Transfer learning improves learning performance by transferring knowledge from different but relevant domains~\cite{Weiss2016_surveytransferlearning}.
In RL for process design, three work~\cite{Gao2023_Transferlearningprocess, Wang2022_Reinforcementlearningautomated, Sachio2022_Integratingprocessdesign} already leveraged transfer learning to accelerate the learning process, e.g., from short-cut simulators to rigorous simulators~\cite{Gao2023_Transferlearningprocess} and from one case study to another case study~\cite{Wang2022_Reinforcementlearningautomated}.
However, in the current transfer learning setting, the agent is still not learning from existing chemical process information. 
Future work can consider leveraging encoder-decoder models such as Variational Autoencoders (VAEs) or transformers to learn from existing flowsheets and then applying transfer learning to the agent.

\subsubsection{Stochastic decision-making}
Considering the uncertainty of energy/feedstock prices and demand is a major challenge for renewable processes~\cite{Mitsos2018_Challengesprocessoptimization}.
However, current RL agents ignore fluctuations in energy/feedstock prices and demand. 
Future research could separate design and operating variables in the RL agent. 
This allows the inclusion of multiple scenarios for flexible operation.
Besides, additional encoders or actors can be included to process stochastic energy prices, demands, and raw material compositions as additional inputs at an operational level.
Therefore, the agent can automatically select the operating variables based on stochastic energy prices and demand in two-stage stochastic programming settings.

\subsubsection{Constrainted decision-making}
Constrained decision-making is crucial for RL in process design to ensure optimal and safe performance.
However, standard RL agent frameworks cannot enforce constraints but include constraints as "soft" penalties in the reward functions~\cite{Stops2022_Flowsheetgenerationhierarchical, Gao2023_Transferlearningprocess, Kalmthout2022_Synthesisseparationprocesses}.
Future work should focus on integrating constraints directly in the agent structure.
As a first step, an additional critic network could be built to account for safety constraints, guiding RL agents to explore appropriate regions in policy optimization~\cite{Yang2022_Safetyconstrainedreinforcement}. 

\subsection{Environment and reward}
In this section, we offer our perspectives on the limitations of the environment and reward setup and provide several suggestions for future work.

\subsubsection{Standardized simulation interfaces }
RL agents frequently interact with process simulators during the training process and the interaction relies on individual interfaces as Table~\ref{tab:literature_table} shows. 
However, current interfaces are usually simulator-specific, which means that a new interface needs to be implemented from scratch whenever a new process simulator is included.
This process is highly repetitive and inefficient, especially for incorporating multi-fidelity process models.
Future work could implement a standardized simulation interface that enables the agent to exchange data efficiently and uniformly between different process simulators.
This interface could potentially make use of existing standards such as CAPE-OPEN~\cite{Jarke1999_CAPEOPENExperiences} and DEXPI+ ~\cite{Theissen2016_DEXPIP&IDSpecification}.

\subsubsection{Multi-fidelity process models}
Current research only leverages a single fidelity model for RL in process design tasks. 
However, RL agents greatly benefit from pre-training on low-fidelity process simulators and subsequent fine-tuning on high-fidelity process simulators~\cite{Gao2023_Transferlearningprocess}. 
Therefore, future research can focus on developing an agent that can dynamically select between multiple fidelity models during training. 
Specifically, a probabilistic model can be developed to guide the RL actor based on multi-fidelity critics to reduce training times and resolve convergence issues.

\subsubsection{Multi-objective rewards}
Current RL frameworks for process design are not suitable for sustainable process design because they are limited to a single objective function.
In the future, the current agent structures could be extended to include multiple objectives.
For example, the critic network could predict multiple rewards, which will be processed by multi-objective optimization to generate corresponding weights for each objective (e.g. economic, environmental, safety)~\cite{Liu2015_MultiobjectiveReinforcementLearning}.

\subsection{Integrated molecular and process design }
\label{2023RLreview:pers_mol_pro}
Current RL frameworks lack the co-design of molecules.
However, the design or selection of molecules is a critical task in many process design tasks, e.g., co-design of working fluids, solvents, or products~\cite{Rehner2023_Moleculesuperstructurescomputer}. 
Also, RL has already been used for molecular design~\cite{Olivecrona2017_Moleculardenovo}.
Therefore, future work should consider integrating these concepts using RL. 
For instance, future work could first use an RL agent for molecular design (e.g., based on~\cite{Olivecrona2017_Moleculardenovo}) to design a solvent. Then, a mechanistic or data-driven model~\cite{Schweidtmann2020_GraphNeuralNetworks} can be used to estimate the relevant properties of the generated molecules. Subsequently, the properties are utilized to simulate the process within the RL for the process design framework. This essentially adds a new hierarchy level to the existing RL for the process design framework.

\subsection{Integrated process operation and design }
\label{2023RLreview:pers_control_design}
Integrating process design and process control becomes increasingly relevant as renewable energy and feedstock demand fluctuates.
For example, the design of a process could be optimized while simultaneously optimizing its operation under changing feedstock compositions or energy prices. 
Another example is the optimal design of batch processes while considering optimal operation strategies. 
These problems are usually formulated as mixed-integer dynamic optimization (MIDO) problems which are difficult to solve.
RL is a tool to solve discrete-time dynamic optimization problems, which makes it suitable for integrating process operation and design.  
However, current RL works only consider process design and operation separately or focus on a specific unit operation design and operation~\cite{Sachio2022_Integratingprocessdesign}.
Future research could integrate process design with process operation through the RL.
For instance, future work could extend the hierarchical RL agent into separate design and operation agents. Then, the design agent defines the design variables, and the operation agent subsequently optimizes operating variables given the current design. Notably, this would require the use of a dynamic simulation environment and will lead to high computational demands. Thus, future research is needed to solve the resulting multi-scale problem efficiently.

\subsection{Benchmarking with established methods}

Numerous established methods are available to solve design optimization problems, including deterministic (e.g., BARON or MAiNGO) and stochastic solvers (e.g., genetic algorithms, Bayesian optimization)~\cite{Boukouvala2016_Globaloptimizationadvances}. 
It is still questionable how RL compares against these traditional approaches for steady-state process design.
Dynamic solution approaches such as RL can in principle be used to solve static optimization problems but are likely significantly less efficient.
However, many process design problems in chemical engineering are actually (mixed-integer) dynamic optimization problems (c.f. Section 3.5). In such instances, RL may be an efficient solution approach. 

Future work should carefully assess the advantages and disadvantages of using RL for steady-state process design and static optimization in general. We recommend conducting comparisons to benchmark different methods. Moreover, we envision the development of new ML-based algorithms that integrate some of the advantages of RL (e.g., large learning capacity and inference capabilities) in the context of process design. For example, encoder-decoder models could be combined with active learning to predict process flowsheet graphs directly~\cite{Hirtreiter2023_automaticgenerationcontrol}.

\section{Conclusions}
We reviewed the state-of-the-art RL in process design in terms of information representation, agent architecture, environment, and reward. 
RL has shown initial promising results for process design but its suitability in static process design still needs to be examined. 
Additionally, a detailed comparison with existing process design methods is missing and current RL frameworks show limited generalization capabilities. 
Therefore, we advocate that future research should benchmark RL with other process design methods.
Additionally, to unlock the full potential of RL, new concepts for meaningful information representation are required. 
Furthermore, the integration of mechanistic knowledge, existing process data, uncertainties, and constraints would be highly beneficial for optimal decision-making.
Finally, future RL frameworks could also integrate molecular design and process operation into the conceptual process design.


\bibliographystyle{unsrt}    
\bibliography{2023_RL_review} 

\end{document}